\documentclass[11pt,a4paper]{article}
\usepackage[hyperref]{acl2019}
\usepackage{times}
\usepackage{latexsym}
\usepackage{url}
\usepackage{graphicx}
\usepackage{subfigure}
\usepackage{amssymb}
\usepackage{amsmath}
\usepackage{paralist,tabularx,multirow}
\usepackage{caption}
\DeclareCaptionFont{10pt}{\fontsize{10pt}{12pt}\selectfont}

\aclfinalcopy 

\title{Joint Effects of Context and User History\\ for Predicting Online Conversation Re-entries}

\author{Xingshan Zeng$^{1,2}$, Jing Li$^3$\thanks{~~~~Jing Li is the corresponding author.}, Lu Wang$^4$, Kam-Fai Wong$^{1,2}$\\
  $^1$The Chinese University of Hong Kong, Hong Kong\\
  $^2$MoE Key Laboratory of High Confidence Software Technologies, China\\
  $^3$Tencent AI Lab, Shenzhen, China\\
  $^4$Northeastern University, Boston, MA, United States \\
  \tt $^{1,2}$\{xszeng,kfwong\}@se.cuhk.edu.hk \\
  \tt $^3$ameliajli@tencent.com, $^4$luwang@ccs.neu.edu \\
}

\date{}

\begin{document}
\maketitle
\begin{abstract}
As the online world continues its exponential growth, interpersonal communication has come to play an increasingly central role in opinion formation and change. 
In order to help users better engage with each other online, we study a challenging problem of re-entry prediction foreseeing whether a user will come back to a conversation they once participated in. We hypothesize that both the context of the ongoing conversations and the users' previous chatting history will affect their continued interests in future engagement. 
%
Specifically, we propose a neural framework with three main layers, each modeling context, user history, and interactions between them, 
to explore how the conversation context and user chatting history jointly result in their re-entry behavior. 
We experiment with two large-scale datasets collected from Twitter and Reddit. Results show that our proposed framework 
with bi-attention 
achieves an F1 score of $61.1$ on Twitter conversations, outperforming the state-of-the-art methods from previous work.
\end{abstract}

\section{Introduction}
\label{sec:intro}

Interpersonal communication plays an important role in information exchange and idea sharing in our daily life. 
We are involved in a wide variety of dialogues every day, ranging from kitchen table conversations to online discussions, all help us make decisions, better understand important social issues, and form personal ideology. 
However, individuals have limited attentions to engage in the massive amounts of online conversations. There thus exists a pressing need to develop automatic conversation management tools to keep track of the discussions one would like to keep engaging in. 
To meet such demand, we study the problem of {\it predicting online conversation re-entries}, where we aim to forecast whether the users will return to a discussion they once entered.

What will draw a user back?
To date, prior efforts for re-entry prediction mainly focus on modeling user’s engagement patterns in the ongoing conversations~\cite{DBLP:conf/wsdm/BackstromKLD13} or rely on the social network structure~\cite{budak2013participation}, largely ignoring the rich information in users' previous chatting history. 

\begin{figure}
\hspace{-2mm}
\includegraphics[width=80mm]{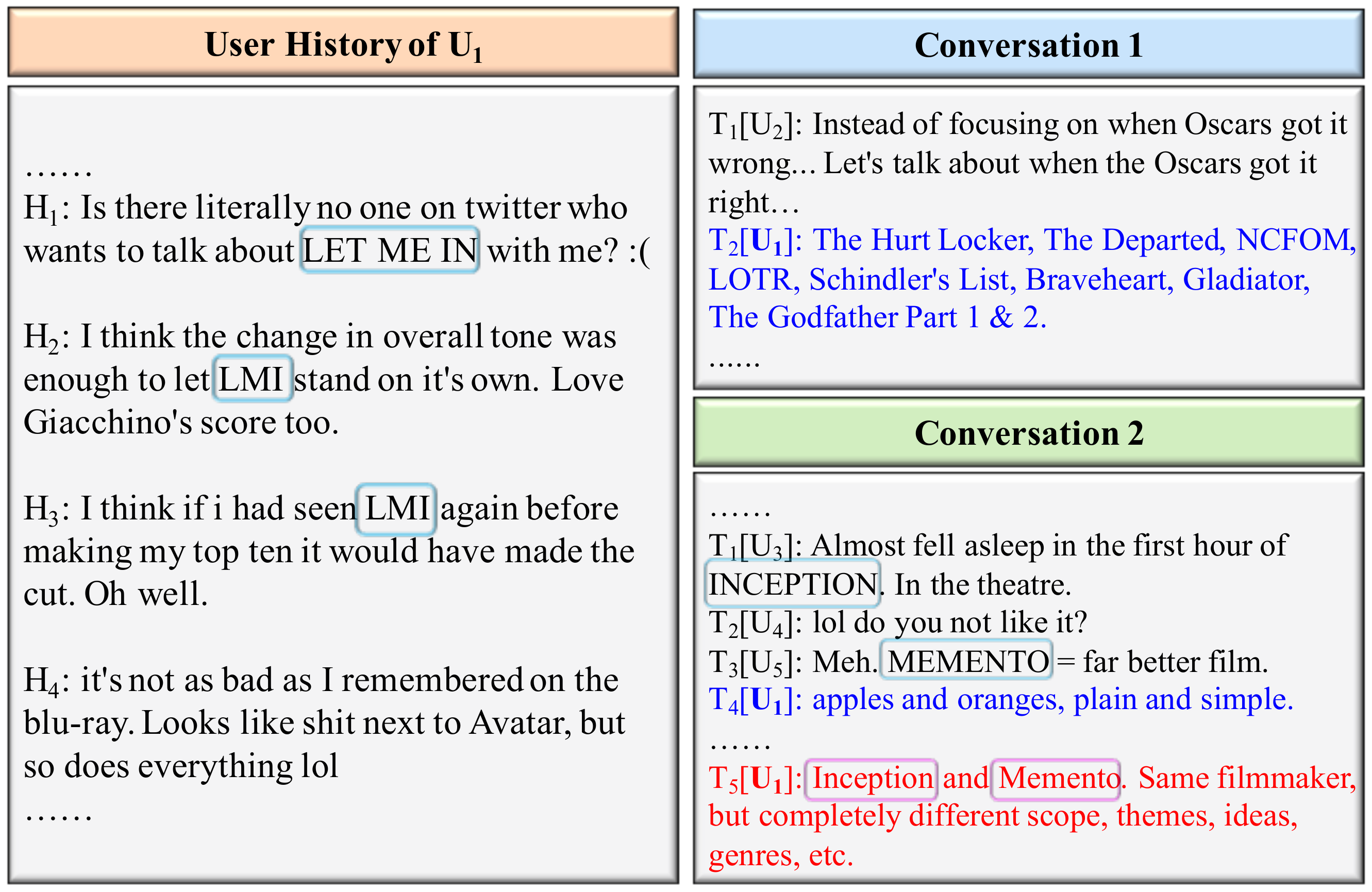}
\captionsetup{font=10pt}
\caption{
Sample tweets in the chatting history of user $\bf U_1$ and two Twitter conversation snippets $\bf U_1$ engaged in.  $\bf H_i$: the $i$-th tweet in $\bf{U_1}$'s history. 
$\bf T_i$[$\bf U_j$]: the $i$-th turn posted by $\bf U_j$. 
First entries by $\bf U_1$ are highlighted in blue in both conversations. $\bf U_1$ only returns to the second one.
}
\vskip -1em 
\label{fig:intro-example}
\end{figure}

Here we argue that effective prediction of one's re-entry behavior requires the understanding of both the \textbf{conversation context}---what has been discussed in the dialogue under consideration, and \textbf{user chatting history} (henceforth user history)---what conversation topics the users are actively involved in. 
In Figure \ref{fig:intro-example}, we illustrate how the two factors together affect a user's re-entry behavior.
Along with two conversations that user $\bf{U_1}$ participated in, also shown is their chatting history in previous discussions. $\bf{U_1}$ comes back to the second conversation since it involves topics on movies (e.g. mentioning \textit{Memento} and \textit{Inception}) and thus suits their interests according to the chatting history, which also talked about movies. 


In this work, we would like to focus on the joint effects of conversation context and user history, ignoring other information. It would be a more challenging yet general task, since information like social networks may be not available in some certain scenarios. 
To study how conversation context and user history jointly affect user re-entries, we propose a novel neural framework that incorporates and aligns the indicative representations from the two information source.
To exploit the joint effects, four mechanisms are employed here:
\emph{simple concatenation} of the two types of representation, \emph{attention} mechanism over turns in context, 
\emph{memory networks}~\cite{sukhbaatar2015end} --- able to learn context attentions in aware of user history,
and \emph{bi-attention} \cite{seo2016bidirectional} --- further capturing interactions from two directions (context to history and history to context). 
More importantly, our framework
enables the re-entry prediction and corresponding representations to be learned in an end-to-end manner. 
On the contrary, previous methods for the same task rely on handcrafted features~\cite{DBLP:conf/wsdm/BackstromKLD13,budak2013participation}, which often require labor-intensive and time-consuming feature engineering processes. 
To the best of our knowledge, we are the first to explore the joint effect of conversation context and user history on predicting re-entry behavior in a neural network framework. 


We experiment with two large-scale datasets, one from Twitter~\cite{DBLP:conf/naacl/ZengLWBSW18}, the other from Reddit which is newly collected\footnote{The datasets and codes are released at: \url{https://github.com/zxshamson/re-entry-prediction}}. Our framework with bi-attention significantly outperforms all the comparing methods including the previous state of the art~\cite{DBLP:conf/wsdm/BackstromKLD13}. For instance, our model achieves an F1 score of $61.1$ on Twitter conversations, compared to an F1 score of $57.0$ produced by \citet{DBLP:conf/wsdm/BackstromKLD13}, which is based on a rich set of handcrafted features. 
Further experiments also show that the model with bi-attention can consistently outperform comparisons given varying lengths of conversation context.
It shows that bi-attention mechanism can well align users' personal interests and conversation context in varying scenarios.

After probing into the proposed neural framework with bi-attention, we find that meaningful representations are learned via exploring the joint effect of conversation context and user history, which explains the effectiveness of our framework in predicting re-entry behavior. 
Finally, we carry out a human study, where we ask two humans to perform on the same task of first re-entry prediction. The model with bi-attention outperforms both humans, suggesting the difficulty of the task as well as the effectiveness of our proposed framework. 


\section{Related Work}


\paragraph{Response Prediction.} 
Previous work on response prediction mainly focuses on predicting whether users will respond to a given social media post or thread. 
Efforts have been made to measure the popularity of a social media post via modeling the response patterns in replies or retweets~\cite{artzi2012predicting,zhang2015retweet}. 
Some studies investigate post recommendation by predicting whether a response will be made by a given user~\cite{chen2012collaborative,yan2012tweet,hong2013co,alawad2016network}.

In addition to post-level prediction, other studies focus on response prediction at the conversation-level. \citet{DBLP:conf/naacl/ZengLWBSW18} investigate microblog conversation recommendation by exploiting latent factors of topics and discourse with a Bayesian model, which often requires domain expertise for customized learning algorithms. 
Our neural framework can automatically acquire the interactions among important components that contribute to the re-entry prediction problem, and can be easily adapted to new domains. 
For the prediction of re-entry behavior in online conversations, previous methods rely on the extraction of manually-crafted features from both the conversation context and the user's social network~\cite{DBLP:conf/wsdm/BackstromKLD13,budak2013participation}. 
Here we tackle a more challenging task, where the re-entries are predicted without using any information from social network structure, which ensures the generalizability of our framework to scenarios where such information is unavailable. 

\paragraph{Online Conversation Behavior Understanding.} 
Our work is also in line with conversational behavior understanding, including how users interact in online discourse~\cite{DBLP:conf/naacl/RitterCD10} and how such behavior signals the future trajectory, including their continued engagement~\cite{DBLP:conf/wsdm/BackstromKLD13,jiao2018find} and the appearance of impolite behavior~\cite{zhang2018conversations}. 
To better understand the structure of conversations, Recurrent Neural Network (RNN)-based methods have been exploited to capture temporal dynamics~\cite{cheng2017factored,DBLP:journals/tacl/ZayatsO18,jiao2018find}. 
Different from the above work, our model not only utilizes the conversations themselves, but also leverages users' prior posts in other discussions.
\section{Neural Re-entry Prediction Combining Context and User History}

\begin{figure}[t]
\includegraphics[width=75mm]{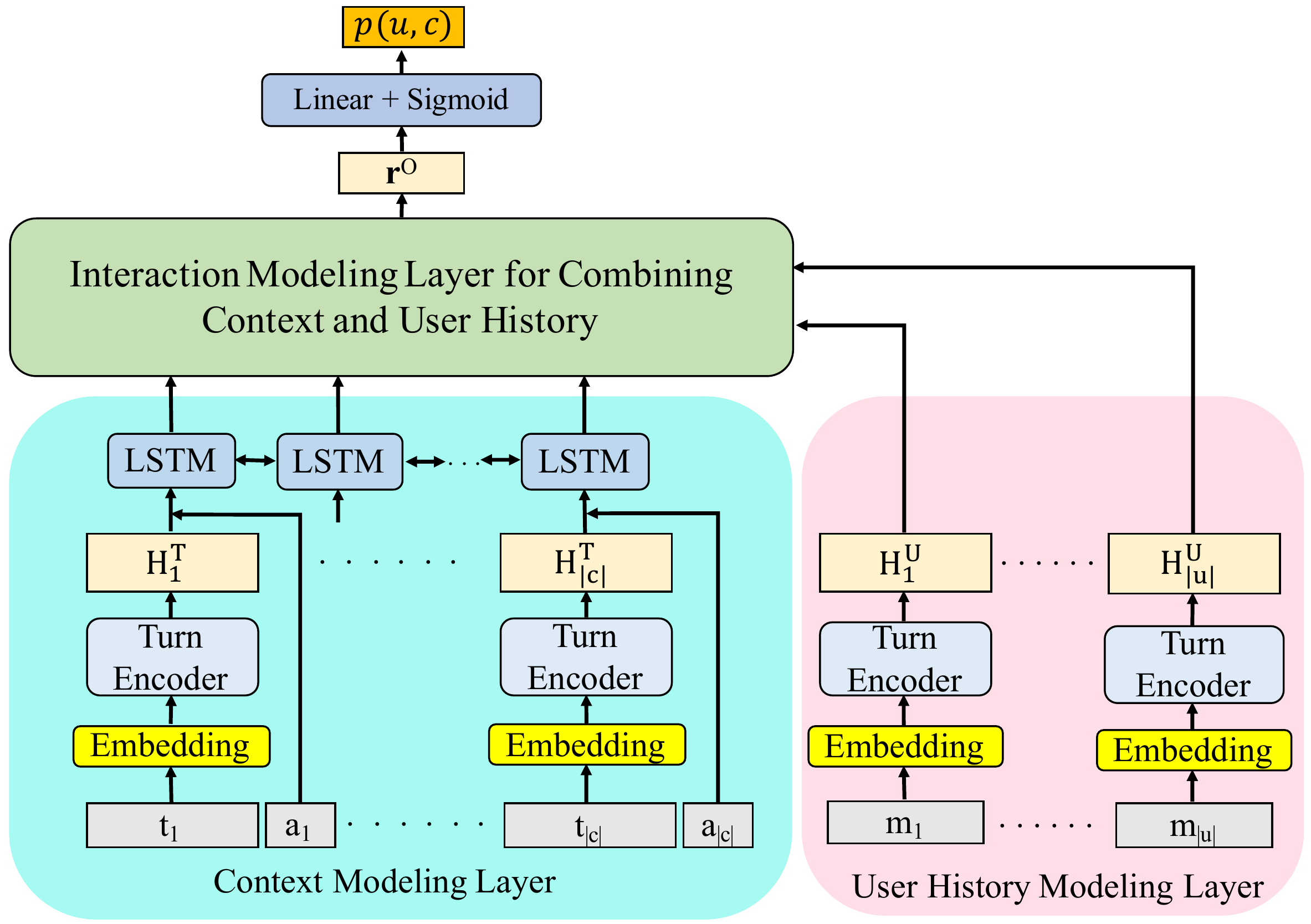}
\captionsetup{font=10pt}
\caption{
The generic framework for re-entry prediction. We implement it with three encoders (Average Embedding, CNN, and BiLSTM) for turn modeling and four mechanisms (Simple Concatenation, Attention, Memory Networks, and Bi-attention) for modeling interactions between context and user history.
}
\label{fig:mod}
\end{figure}

This section describes our neural network-based conversation re-entry prediction framework exploring the joint effects of context and user history.
Figure \ref{fig:mod} shows the overall architecture of our framework, consisting of three main layers: context modeling layer, user history modeling layer, and interaction modeling layer to learn how information captured by the previous two layers interact with each other and make decisions conditioned on their joint effects.
Here we adopt four mechanisms for interaction modeling: simple concatenation, attention, memory networks, and bi-attention, which will be described later. 




\subsection{Input and Output}\label{ssec:model:io}
We start with formulating model input and output. At input layer, our model is fed with two types of information, the chatting history of the target user $u$ and the observed context of the target conversation $c$. 
The goal of our model is to output a Bernoulli distribution $p(u,c)$ indicating the estimated likelihood of whether $u$ will re-engage in the conversation $c$. Below gives more details.

Formally, we formulate the context of $c$ as a sequence of chronologically ordered turns $\langle t_1,t_2,\cdots,t_{|c|}\rangle$, where the last turn $t_{|c|}$ is posted by $u$ (we then predict $u$'s re-entries afterwards). 
Each turn $t$ is represented by a sequence of words ${\bf w}_t$, and an auxiliary triple, ${\bf a}_t=\langle i_{t}, r_{t}, u_{t}\rangle$, 
%
where $i_{t}$, $r_{t}$, and $u_{t}$ are three indexes indicating the position of turn $t$, which turn $t$ replies to, and the author of $t$, respectively. 
Here ${\bf a}_{t}$ is used to record the replying structures as well as the user's involvement pattern. 

For the user history, we formulate it as a collection of $u$'s chatting messages $\{m_1, m_2, \cdots, m_{|u|}\}$, all posted before the time $t_{|c|}$ occurs.
Each message $m$ is denoted as its word sequence, ${\bf w}_m$. 

In the following, we explain how the aforementioned representations are processed by our model to make predictions. 
The three main layers in Figure~\ref{fig:mod} are described in Sections \ref{ssec:model:context}, \ref{ssec:model:user}, and \ref{ssec:model:bi-attention}, respectively. The learning objective is presented in Section \ref{ssec:model:final}.

\subsection{Context Modeling Layer}\label{ssec:model:context}

The context modeling layer captures representations from the observed context for the target conversation $c$. 
To this end, we jointly model the content in each turn (henceforth \textbf{turn modeling}) and the turn interactions in conversation structure (henceforth \textbf{structure modeling}). 

\paragraph{Turn Modeling.} 
The turn representations are modeled via turn-level word sequence with a turn encoder. 
We exploit three encoders here: \textbf{Average Embedding} (Averaging each word's embedding representation), \textbf{CNN} (Convolutional Neural Networks), and \textbf{BiLSTM} (Bidirectional Long Short-Term Memory).
BiLSTM's empirical performance turns out to be slightly better (will be reported in Table~\ref{tab:main_res}).

Concretely, given the conversation turn $t$, each word $w_i$ of $t$ is represented as a vector mapped by an embedding layer $I(\cdot)$, which is initialized by pre-trained embeddings and updated during training. 
The embedded vector $I(w_i)$ is then fed into the turn encoder, yielding the \textit{turn representation} for $t$, denoted by $H^T_{t}$.\footnote{For all the BiLSTM encoders in this work, without otherwise specified, we take the concatenation of all hidden states from both the directions as its learned representations.}

\paragraph{Structure Modeling.} 
To learn the conversational structure representations for $c$, our model applies BiLSTM, namely structure encoder, to capture the interactions between adjacent turns in its context.  
Each state of this structure encoder sequentially takes $t$'s turn representation, $H_{t}^T$, concatenated with the auxiliary triple, ${\bf a}_t$, as input to produce the structure representation $H^C$. 
Our intuition is that $H^C$ should capture both the content of the conversation and interaction patterns among its participants.  
Then $H^C$, considered as the \textit{context representation} for $c$, is sent to interaction modeling layer as part of its input. 

\subsection{User History Modeling Layer}\label{ssec:model:user} 
To encode the user history for target user $u$, in this layer, we first apply the same encoder in turn modeling to encode each chatting message $m$ by $u$, as they both explore the post-level representations.
The turn encoder is sequentially fed with the embedded word in $m$, and produce the message-level representation $H^M_m$. All messages in $u$'s user history are further concatenated into a matrix $H^U$, serving as $u$'s \textit{user history representation} and the input of the next layer.

\subsection{Interaction Modeling Layer}\label{ssec:model:bi-attention}

To capture whether the discussion points in $c$ match the interests of $u$,  
$H^C$ (from context modeling) and $H^U$ (from user history modeling) are merged through an interaction modeling mechanism over the two sources of information.
We hypothesize that users will be likely to come back to a conversation if its topic fits their own interests. 
Here, we explore four different mechanisms for interaction modeling.
Their learned interaction representation, denoted as ${\bf r}^O$, is fed into a sigmoid-activated neural perceptron~\cite{glorot2011deep}, for predicting final output $p(u,c)$. 
It indicates how likely the target user $u$ will re-engage in the target conversation $c$. 
We then describe the four mechanisms to learn ${\bf r}^O$ in turn below.


\paragraph{Simple Concatenation.} 
Here we simply put context representation (last state) and user representations (with average pooling) side by side, yielding ${\bf r}^O=[H^C_{|c|}; \sum_j^{|u|} H^U_j / |u|]$ as the interaction representation for re-entry prediction.

\paragraph{Attention.} 
To capture the context information useful for re-entry prediction, we exploit an attention mechanism~\cite{luong2015effective} over $H^C$.
Attentions are employed to ``soft-address'' important context turns according to their similarity with user representation (with average pooling). 
Here we adopt dot attention weights and define the attended interaction representation as:
\vspace{-1mm}
\begin{equation}\small
    {\bf r}^O= \sum_i^{|c|} \alpha_{i}\cdot H^C_i\text{, } \alpha_{i}=softmax(H_i^C\cdot \sum_j^{|u|} H^U_j/{|u|})
    \vspace{-1mm}
\end{equation}

\paragraph{Memory Networks.}

To further recognize indicative chatting messages in user history, we also apply end-to-end memory networks (MemN2N)~\cite{sukhbaatar2015end} for interaction modeling.
It can be seen as a recurrent attention mechanism over chatting messages (stored in memory).
Hence fed with context representation, memory networks will yield a memory-aware vector as interaction representation:
\vspace{-2mm}
\begin{equation}\small \label{mem}
    {\bf r}^O= \sum_j^{|u|} \alpha_{j}\cdot f_{turn}(H^U_j)\text{, } \alpha_{j}=softmax(H^C_{|c|}\cdot H^U_j )
    \vspace{-2mm}
\end{equation}
\noindent where $f_{turn}(\cdot)$ denotes the unit function used for turn modeling.

Here we adopt multi-hop memory mechanism to allow deep user interests to be learned from chatting history. For more details, we refer the readers to \citet{sukhbaatar2015end}. 

\paragraph{Bi-attention.}
Inspired by~\citet{seo2016bidirectional}, we also apply bi-attention mechanism to explore the joint effects of context and user history.
Intuitively, the bi-attention mechanism looks for evidence, if any, indicating the topics of the current conversation that align with the user's interests from two directions (i.e. context to history and history to context), such as the names of two movies \textit{Inception} and \textit{Let Me In} shown in Figure \ref{fig:intro-example}. Concretely, bi-attention mechanism captures context-aware attention over user history messages:
\begin{equation}\small
\alpha^U_{ij}=\frac{\exp(f_{score}(H^C_i,H^U_j))}{\sum_{j'=1}^{|u|} \exp(f_{score}(H^C_i,H^U_{j'}))}
\end{equation}
\noindent where the alignment score function takes a form of $f_{score}(H^C_i,H^U_j)= W_{bi-att}[H^C_i;H^U_j;H^C_i \circ H^U_j]$. 
It captures the similarity of the $i$-th context turn and the $j$-th user history message. The weight vector $W_{bi-att}$ is learnable in training.

Likewise, we compute user-aware attention over context turns.
Afterwards, the bi-directional attended representations are concatenated and passed into a ReLU-activated multilayer perceptron (MLP), yielding representation $\bf r$.
$\bf r$, as turn-level representation, is then sequentially fed into a two-layer BiLSTM, to produce the interaction representation ${\bf r}^O$.

\subsection{Learning Objective}\label{ssec:model:final}

For parameter learning in our model, we design the objective function based on cross-entropy loss as following:

\vspace{-2mm}
\begin{equation} \label{loss}
\small
\mathcal{L} = - \sum_i \big [ \lambda y_i \log (\hat{y}_i) + \mu (1 - y_i) \log (1 - \hat{y}_i) \big ]
\vspace{-2mm}
\end{equation}
\noindent where the two terms reflect the prediction on positive and negative instances, respectively.
Moreover, to take the potential data imbalance into account, we adopt two trade-off weights $\lambda$ and $\mu$. 
The parameter values are set based on the proportion of positive and negative instances in the training set (see Section~\ref{sec:expsetup}). 
$\hat{y}_i$ denotes the re-entry probability estimated from $p(u,c)$ for the $i$-th instance, 
and $y_i$ is the corresponding binary ground-truth label ($1$ for re-entry and $0$ for the opposite).
\section{Experimental Setup}
\label{sec:expsetup}
\paragraph{Data Collection and Statistic Analysis.} 
To study re-entry behavior in online conversations, we collected two datasets: one is released by \citet{DBLP:conf/naacl/ZengLWBSW18} containing Twitter conversations formed by tweets from the TREC 2011 microblog track data\footnote{\url{https://trec.nist.gov/data/tweets/}} (henceforth \textbf{Twitter}), and the other is \textit{newly collected} from Reddit (henceforth \textbf{Reddit}), a popular online forum. In our datasets, the conversations from Twitter concern diverse topics, while those from Reddit focus on the political issues. Both datasets are in English.

To build the Reddit dataset, we first downloaded a large corpus publicly available on Reddit platform.\footnote{\url{https://www.reddit.com/r/datasets/comments/3bxlg7/i_have_every_publicly_ available_reddit_comment/}} Then, we selected posts and comments in subreddit ``politics'' posted from Jan to Dec 2008. Next, we formed Reddit posts and comments into conversations with replying relations revealed by the ``\textit{parent\_id}'' of each comment. Last, we removed conversations with only one turn.

In our main experiment, we focus on {\it first re-entry} prediction, i.e. we predict whether a user $u$ will come back to a conversation $c$, given current turns until $u$'s first entry in $c$ as context and $u$'s past chatting messages (posted before $u$ engaging in $c$). 
For model training and evaluation, we randomly select $80$\%, $10$\%, and $10$\% conversations to form training, development, and test sets.


\begin{table}[t]
\begin{center}
\setlength{\tabcolsep}{1.0mm}
\fontsize{10}{11}\selectfont
\begin{tabular}{|l|r r|}
\hline 
&\bf{Twitter}&\bf{Reddit}\\
\hline
\# of users&10,122&13,134\\
\# of conversations&7,500&29,477\\
\# of re-entry instances&5,875&12,780\\
\# of non re-entry instances &8,677&39,988\\
\hline
Avg. \# of convs per user&1.7&5.9\\
Avg. \# of msgs in user history &3.9&8.4\\
Avg. \# of entries per user per conv&2.0&1.3\\
Avg. \# of turns per conv&5.2&3.7\\
Avg. \# of users per conv&2.3&2.6\\

\hline
\end{tabular} 
\end{center}
\captionsetup{font=10pt}
\caption{Statistics of two datasets.
}
\label{tab:stat}
\end{table}

The statistics of the two datasets are shown in Table \ref{tab:stat}. As can be seen, users participate twice on average in Twitter conversations, and the number is only $1.3$ on Reddit. This results in the severe imbalance over instances of re-entry and non re-entry (negative samples where users do not come back) on both datasets. Therefore, strategies should be adopted for alleviating the data imbalance issue, as done in Eq. (\ref{loss}). 
It indicates the sparse user activity in conversations, where most users engage in a conversation only once or twice.
Thus predicting user re-entries only with context will not perform well, and the complementary information underlying user history should be leveraged.

\begin{figure}[t]
\centering
\subfigure[User history]{
\includegraphics[width=0.22\textwidth]{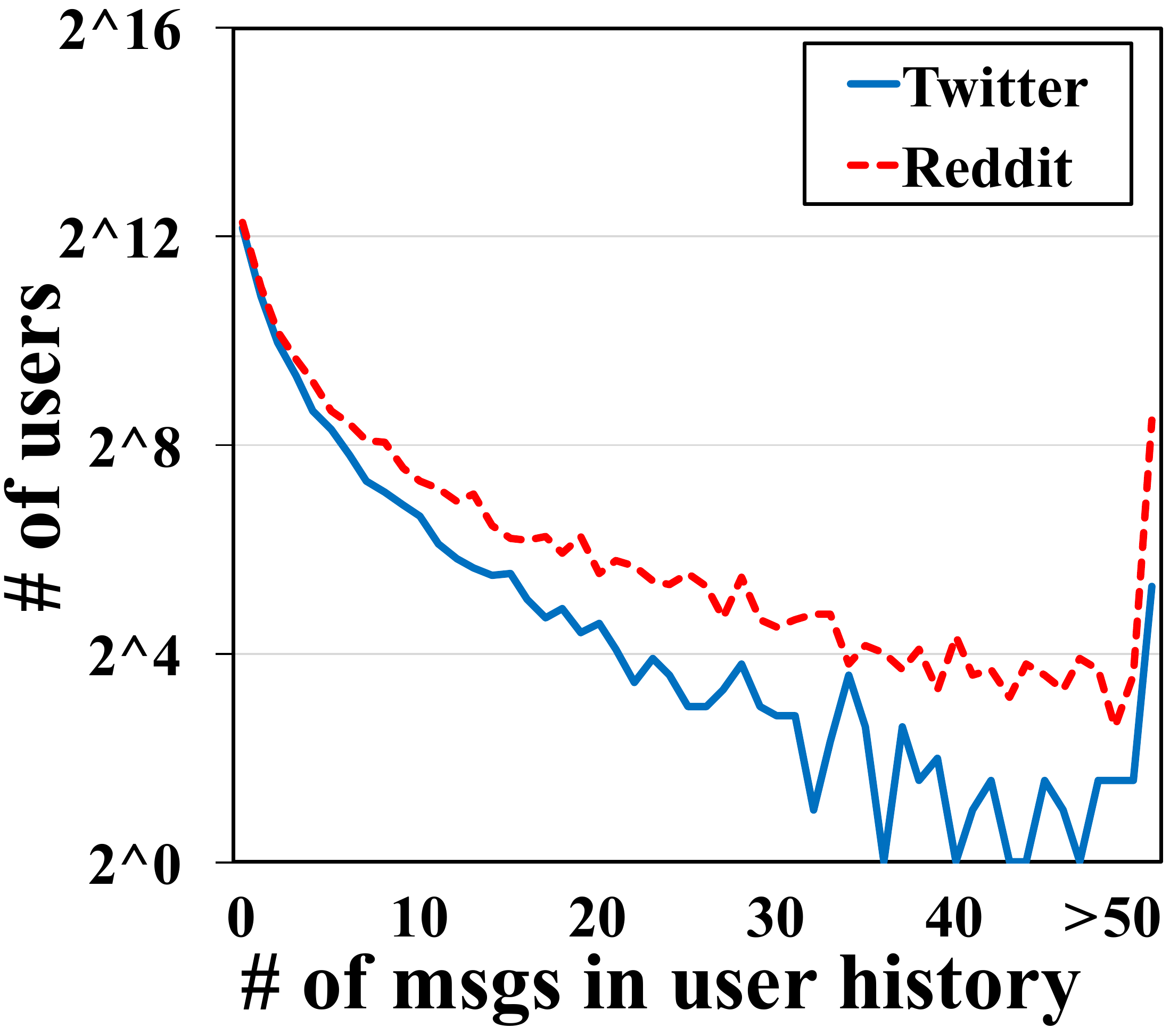}\label{subfig:user-history}
}
\subfigure[Conversation context]{
\includegraphics[width=0.225\textwidth]{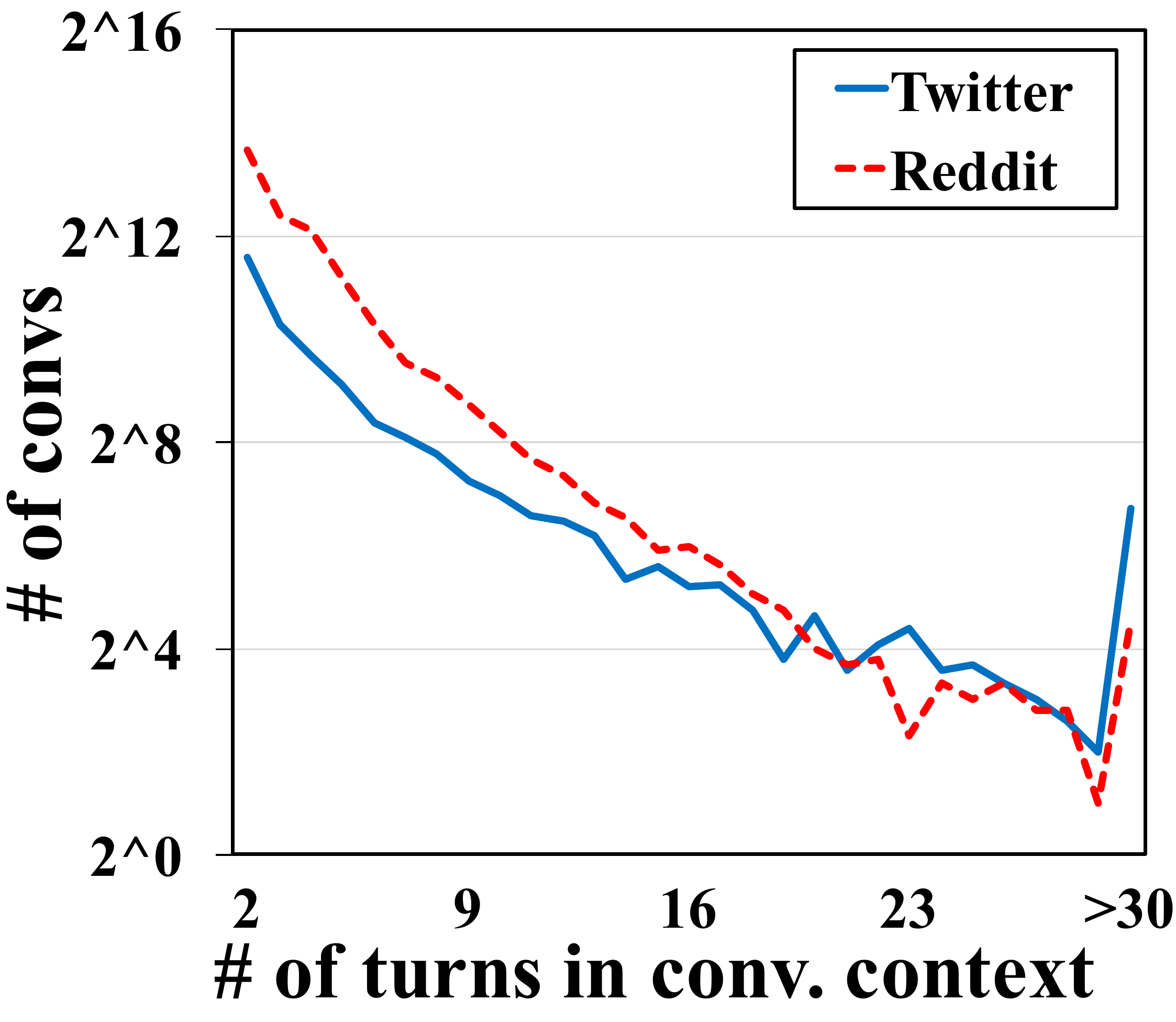}\label{subfig:conv-context}
}
\captionsetup{font=10pt}
\caption{
Distributions of message number in user history and turn number in conversation context on the two datasets. 
}
\label{fig:distribution}
\end{figure}

We further study the distributions of message number in user history and turn number in conversation context on both datasets. 
As shown in Figure \ref{fig:distribution}, 
there exists severe sparsity in either user history or conversation context.
Thus combining them both might help alleviate the sparsity in one information source.
We also notice that Twitter and Reddit users exhibit different conversation behaviors. Reddit users tend to engage in more conversations, resulting in more messages in user history (as shown in Figure \ref{subfig:user-history}). Twitter users are more likely to stay within each conversation, leading to lengthy discussions and larger re-entry frequencies on average, as shown in Figure \ref{subfig:conv-context} and Table \ref{tab:stat}.

\paragraph{Data Preprocessing and Model Setting.}

For preprocessing Twitter data, we applied Glove tweet preprocessing toolkit~\cite{pennington2014glove}.\footnote{\url{https://nlp.stanford.edu/projects/glove/preprocess-twitter.rb}}
For the Reddit dataset, we first applied the open source natural language toolkit (NLTK)~\cite{loper2002nltk} for word tokenization.
Then, we replaced links with the generic tag ``URL'' and removed all the non-alphabetic tokens. 
For both datasets, a vocabulary was built and maintained in experiments with all the tokens (including emoticons and punctuation) from training data.

For model setups, we initialize the embedding layer with $200$-dimensional Glove embedding~\cite{pennington2014glove}, where Twitter version is used for our Twitter dataset and the Common Crawl version applied on Reddit dataset.\footnote{\url{https://nlp.stanford.edu/projects/glove/}}
All the hyper-parameters are tuned on the development set by grid search.
The batch size is set to $32$.
Adam optimizer \cite{kingma2014adam} is adopted for parameter learning with initial learning rate selected among $\{10^{-3},10^{-4},10^{-5}\}$. 
For the BiLSTM encoders, we set the size of their hidden states to $200$ ($100$ for each direction).
For the CNN encoders, we use filter windows of $2$, $3$, and $4$, each with $50$ feature maps.
In MemN2N interaction mechanism, we set hop numbers to $3$.
In the learning loss, we set $\mu=1$ and $\lambda=2$, the weights to tackle data imbalance.
For re-entry prediction, a user is considered to come back if the estimated probability for re-entry is larger than $0.5$.

\paragraph{Baselines and Comparisons.} 
For comparisons, we consider three baselines. 
\textsc{Random} baseline: randomly pick up a ``yes-or-no'' answer.
\textsc{History} baseline: predict based on users' history re-entry rate before current conversation, which will answer ``yes'' if the rate exceeds a pre-defined threshold (set on development data), and ``no'' otherwise. 
(For users who lack such information before current conversation, it predicts ``yes or no'' randomly.)
\textsc{All-Yes} baseline: always answers ``yes'' in re-entry prediction. 
Its assumption is that users tend to be drawn back to the conversations they once participated by the platform's auto messages inviting them to return.

For supervised models, we compare with \textsc{CCCT}, the state-of-the-art method proposed by \citet{DBLP:conf/wsdm/BackstromKLD13}, where the bagged decision tree with manually-crafted features (including arrival patterns, timing effects, most related terms, etc.) are employed for re-entry prediction. We do not compare with \citet{budak2013participation}, since most of its features are related to social networks or Twitter group information, which is unavailable in our data. 

In our proposed neural framework, we further compare varying encoders for turn modeling and mechanisms to model the interactions between user history and conversation context.
We first compare three turn encoders --- \textsc{Avg-Embed} (average embedding), \textsc{CNN}, and \textsc{BiLSTM}, to examine their performance in turn representation learning. 
Their results are compared on our variant only with context modeling layer and the best encoder (turned out to be \textsc{BiLSTM}) is applied on the full model.
For the interaction modeling layer, we also study the effectiveness of  four mechanisms to combine user history and conversation context --- simple concatenation (\textsc{Con}), attention (\textsc{Att}), memory networks (\textsc{Mem}), and bi-attention (\textsc{BiA}). 

\section{Results and Analysis}
This section first discusses prediction results of first re-entry in Section~\ref{ssec:first-reentry}. 
We then present the results of the second and third re-entry prediction in Section \ref{ssec:context-and-history}, as well as an analysis on user history effects.
Section~\ref{ssec:discussion} then provides explanations on what we learn from the joint effects from context and user history, indicative of user re-entries.
Finally, we conduct a human study to compare human performance on the same task with our best model (Section~\ref{ssec:human}).

\subsection{First Re-entry Prediction Results}\label{ssec:first-reentry}

\begin{table*}
\begin{center}
\fontsize{11}{11}\selectfont
\begin{tabular}{|l|cccr|cccr|}
\hline
\multirow{2}{*}{\bf{Models}}&\multicolumn{4}{c|}{\bf Twitter}& \multicolumn{4}{c|}{\bf Reddit}\\
\cline{2-9}
&AUC&F1 Score&Precision&Recall&AUC&F1 Score&Precision&Recall\\
\hline
\underline{\bf Baselines} & & & & & & & & \\
\textsc{Random}&51.0&45.0&40.3&50.9&49.4&32.6&24.5&48.7 \\
\textsc{History}&50.1&46.4&42.2&51.4&50.7&35.2&26.9&50.9\\
\textsc{All-Yes}&50.0&54.9&37.9&\bf{100.0}&50.0&38.5&23.8&\bf{100.0}\\
\hline
\hline
\underline{\bf S.O.T.A} & & & & & & & & \\
\textsc{CCCT}&57.7&57.0&45.5&76.4&59.9&39.8&\bf{44.7}&36.0 \\
\hline
\hline
\underline{\bf W/O History} & & & & & & & & \\
\textsc{Avg-Embed}&60.4&59.0&43.5&91.8&63.7&42.4&31.0&67.2\\
\textsc{CNN}&58.8&59.1&43.2&93.5&64.0&42.8&31.1&68.5\\
\textsc{BiLSTM}&60.4&59.4&45.8&85.0&64.1&43.1&31.4&69.5\\
\hline
\hline
\underline{\bf With History} & & & & & & & & \\
\textsc{BiLSTM+Con}&51.0&58.0&40.9&\bf{100.0}&50.1&38.6&24.0&98.3\\
\textsc{BiLSTM+Att}&58.4&59.0&44.6&87.3&60.3&41.3&27.8&82.4\\
\textsc{BiLSTM+Mem}&61.3&59.9&45.7&87.5&65.5&43.7&31.8&69.9\\
\textsc{BiLSTM+BiA}&\bf{62.7}&\bf{61.1}&\bf{47.0}&87.7&\bf{67.1}&\bf{45.4}&33.9&68.9\\ 
\hline
\end{tabular}
\end{center}
\captionsetup{font=10pt}
\caption{Results on first re-entry prediction. The best results in each column are in \textbf{bold}. Model \textsc{BiLSTM+BiA} yields significantly better AUC and F1 scores than all other comparisons ($p<0.05$, paired t-test). 
}
\label{tab:main_res}
\end{table*}

In main experiment, we adopt the automatic evaluation metrics --- AUC, F1 score, precision, and recall, and focus on the prediction of the major re-entry type --- {\it first re-entry}, where conversation context up to user's first participation is given. 
As shown in Table \ref{tab:stat}, most users, if re-entry, only return once to a conversation. Also, in  conversation management, the prediction of first re-entry is a challenging yet practical problem.
We will discuss second and third re-entry prediction later in Section \ref{ssec:context-and-history}.
The comparison results 
are reported in Table \ref{tab:main_res}. 
On both datasets, we observe: 

\vspace{0.5em}

$\bullet$~\textit{First re-entry prediction is challenging}. All models produce AUC and F1 scores below $70$. In particular, models built on rules and features with shallow content and network features perform poorly, suggesting the need of better understanding of conversations or more information like user's chatting history. 
We also observe that \textsc{History} yields only slightly better results than \textsc{Random}.
It suggests that users' re-entries depend on not only their past re-entry patterns, but also the conversation context.


$\bullet$~\textit{Well-encoded user chatting history is effective.} 
Among neural models, our \textsc{BiLSTM+Mem} and \textsc{BiLSTM+BiA} models outperform other comparisons by successfully modeling users' previous messages and their alignment with the topics of ongoing conversations. 
However, the opposite observation is drawn for \textsc{BiLSTM+Con} and \textsc{BiLSTM+Att}.
It is because the interactions between context and user history are effective yet complex, requiring well-designed merging mechanisms to exploit their joint effects.

$\bullet$~\textit{Bi-attention mechanism better aligns the users' interests and the conversation topics.} 
\textsc{BiLSTM+BiA} achieves the best AUC and F1 scores, significantly outperforming all other comparison models on both datasets. In particular, it beats \textsc{BiLSTM+Mem}, which also able to learn the interaction between user history and conversation content, indicating the effectiveness of bi-attention over memory networks in this task. 

Interestingly, comparing the results on the two datasets, we notice all models yield better recall and F1 on Twitter than Reddit. 
This is due to the fact that Reddit users are more likely to abandon conversations, reflected as the fewer number of entries in Table \ref{tab:stat}. Twitter users, on the other hand, tend to stay longer in the conversations, which encourages all models to predict the return of users. 

\begin{figure}[t]
\centering
\subfigure[Twitter Dataset]{
\includegraphics[width=0.223\textwidth]{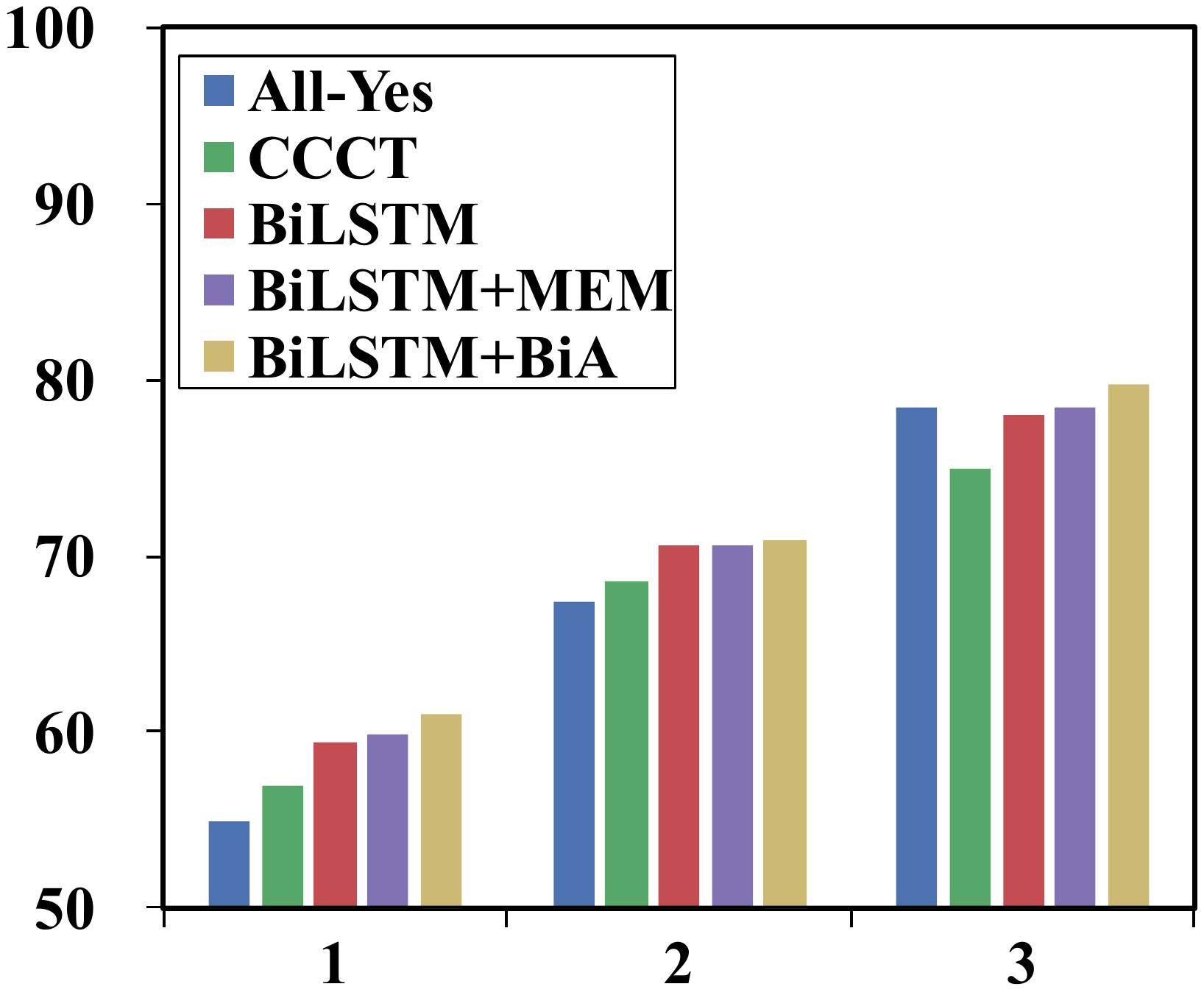}
}
\subfigure[Reddit Dataset]{
\includegraphics[width=0.22\textwidth]{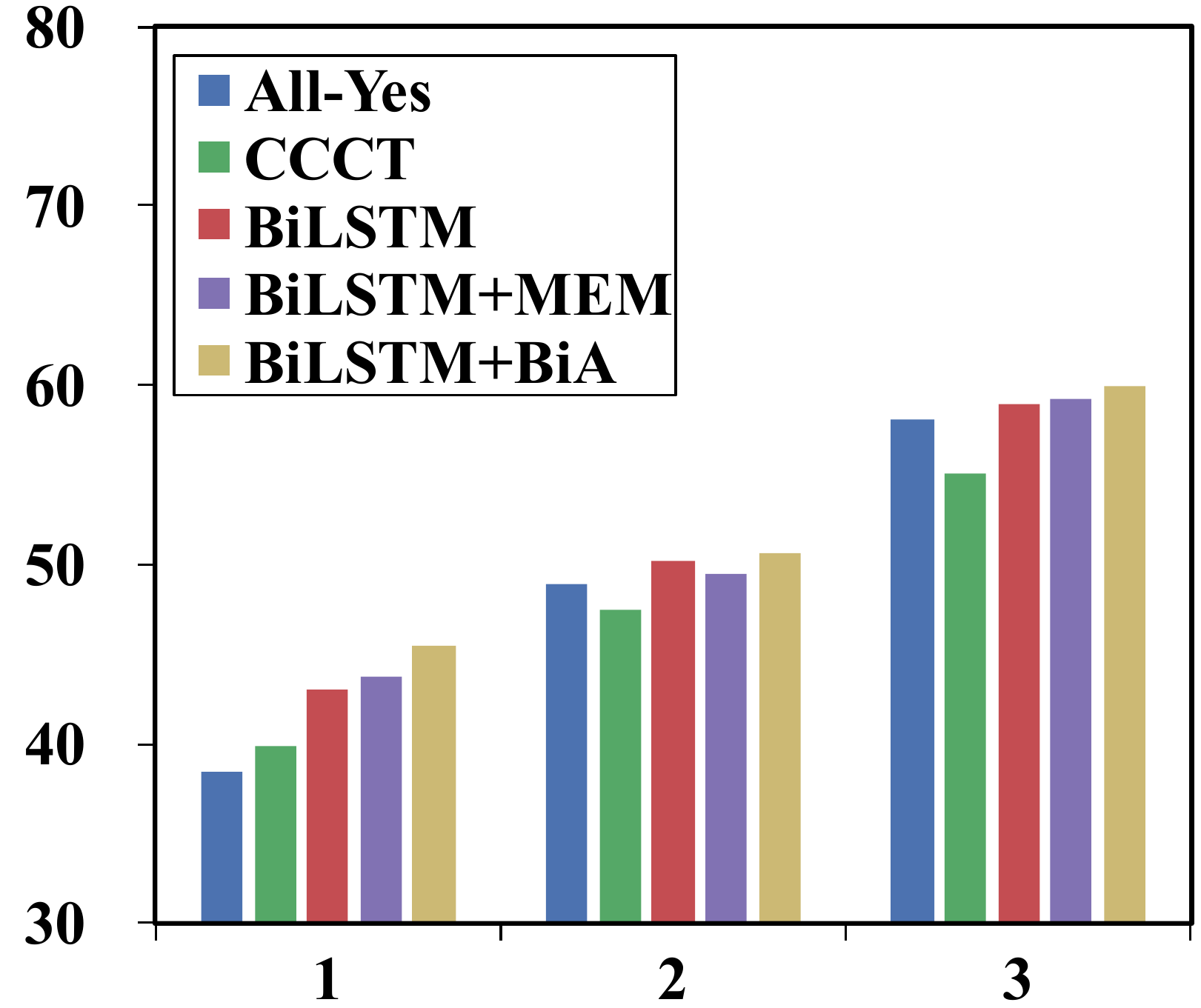}
}
\captionsetup{font=10pt}
\caption{
F1 scores for prediction on the first, second, and third re-entries (given the conversation context until the last entry). 
X-axis: \# of turns in the given conversation context.
Both figures, from left to right, show the F1 scores by \textsc{All-Yes}, \textsc{CCCT}, \textsc{BiLSTM}, \textsc{BiLSTM+MEM}, and \textsc{BiLSTM+BiA}. 
}
\label{fig:diff_times}
\end{figure}



\subsection{Predicting Re-entries with Varying Context and User History}\label{ssec:context-and-history}

Here we study the effects of varying conversation context and user history over re-entry prediction.

\paragraph{Results with Varying Context.}

We first discuss model performance given different amounts of conversation context by varying the number of user entries. 
Figure \ref{fig:diff_times} shows the F1 scores for predicting the first, second, and third re-entries. For predicting second or third re-entries, turns of current context until given user's second or third entry will be given. 
As can be seen, all models' performance monotonically increases when more context is observed. Our \textsc{BiLSTM+BiA} uniformly outperforms other methods in all setups. 
Interestingly, baseline \textsc{All-Yes} achieves the most performance gain when additional context is given. This implies that the more a user contributes to a conversation, the more likely they will come back. 


\paragraph{Results with Varying User History.}

\begin{figure}[t]
\setlength{\belowcaptionskip}{-4mm}
\centering
\includegraphics[width=0.40\textwidth]{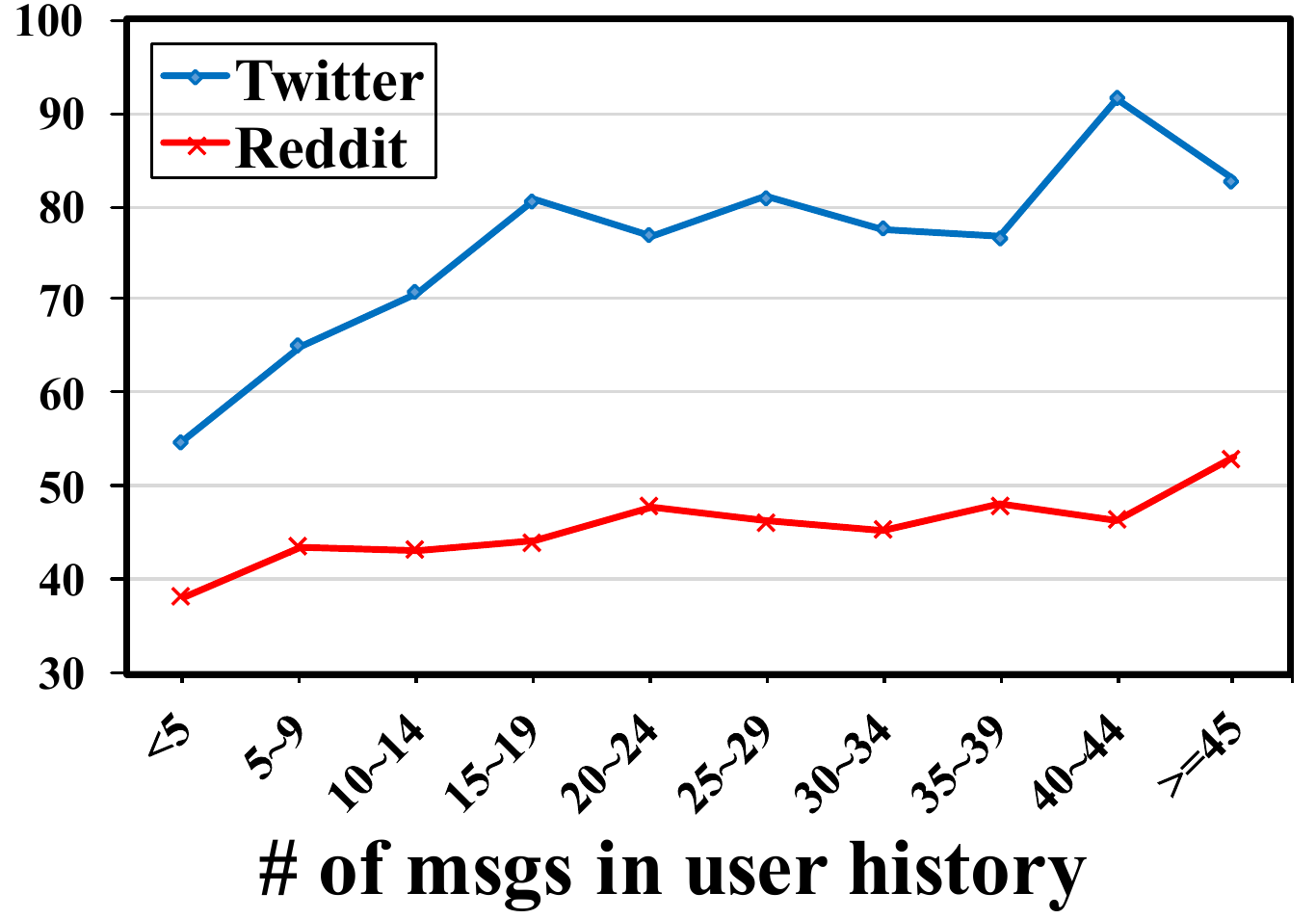}
\captionsetup{font=10pt}
\caption{ 
F1 scores of model \textsc{BiLSTM+BiA} on first re-entry prediction, with varying numbers of chatting messages given in user history. 
}
\label{fig:diff_len}
\end{figure}

We further analyze how model performance differs when different amounts of messages are given in the user history. From Figure \ref{fig:diff_len}, we can see that it generally yields better F1 scores when more messages are available for the user history, suggesting the usefulness of chatting history to signal user re-entries. The performance on Reddit does not increase as fast as observed on Twitter, which may mainly because the context from Reddit conversations is often limited. 


\subsection{Further Discussion}\label{ssec:discussion}

We further discuss our models with an ablation study and a case study to understand and interpret their prediction results.

\paragraph{Ablation Study.} 
To examine the contribution of each component in our framework, we present an ablation study on first re-entry prediction task.
Table \ref{tab:ablation} shows the results of our best full model (\textsc{BiLSTM+BiA}) together with its variant without using turn-level auxiliary meta ${\bf a}_t$ (defined in Section \ref{ssec:model:io} to record user activity and replying relations in context), and that without structure modeling layer (to capture conversation discourse in context described in Section \ref{ssec:model:context}); also compared are variants without using user chatting history (described in Section \ref{ssec:model:user}).

\begin{table}[th]
\begin{center}
\setlength{\tabcolsep}{1.3mm}
\fontsize{10}{11}\selectfont
\begin{tabular}{|l|lll|lll|}
\hline 
\multirow{2}{*}{\bf{Models}}&\multicolumn{3}{c|}{\bf Twitter}& \multicolumn{3}{c|}{\bf Reddit}\\
\cline{2-7}
&F1&Pre&Rec&F1&Pre&Rec\\
\hline
\underline{\bf W/O History} & & & & & & \\
W/O SML&58.8&42.6&\textbf{95.1}&39.6&25.2&\textbf{92.9}\\
With SML&59.4&45.9&85.0&43.1&31.4&69.5\\
\hline
\underline{\bf With History} & & & & & & \\
W/O SML&57.5&43.2&86.7&43.8&31.3&74.4\\
W/O Meta&60.4&46.6&86.1&44.3&31.3&75.8\\
Full model&\textbf{61.1}&\textbf{47.0}&87.7&\textbf{45.4}&\textbf{33.9}&68.9\\
\hline
\end{tabular} 
\end{center}
\captionsetup{font=10pt}
\caption{Results of our variants.
SML: structure modeling layer. Meta: auxiliary triples ${\bf a}_t$.
Our full model \textsc{BiLSTM+BiA} obtains the best F1. 
}
\label{tab:ablation}
\end{table}

Our full model yields the best F1 scores, showing the joint effects of context and user history can usefully indicate user re-entries.
We also see that auxiliary triples, though conveying simple meta data for context turns, are helpful in our task.
In addition, interestingly, conversation structure looks more effective in models leveraging user history, because they can learn deeper semantic relations between context turns and user chatting messages.



\paragraph{Case Study.} 
We further utilize a case study based on the sample conversations shown in Figure~\ref{fig:intro-example} to demonstrate what our model learns.
Table~\ref{tab:caseres} displays the outputs from different models on estimating how likely $\bf{U_1}$ will re-engage in conversation 1 ($\bf{C_1}$) and conversation 2 ($\bf{C_2}$), where $\bf{U_1}$ returns to the latter. 
All neural models successfully forecast that $\bf{U_1}$ is more likely to re-engage in $\bf{C_2}$, while only \textsc{BiLSTM+BiA} yields correct results (given threshold $0.5$). 


\begin{table}[t]
\centering
\fontsize{10}{11}\selectfont
\begin{tabular}{|l|c|c|}
\hline 
\bf{Models}& Conv. 1 ($\bf{C_1}$)& Conv. 2 ($\bf{C_2}$)\\
\hline
\textsc{CCCT}&1.0&1.0\\
\textsc{BiLSTM}&0.386&0.480\\
\textsc{BiLSTM+Mem}&0.583&0.712\\
\textsc{BiLSTM+BiA}&0.460&0.581\\
\hline
\end{tabular} 
\captionsetup{font=10pt}
\caption{Predicted probabilities by different models for user ${\bf U_1}$'s re-entry to conversations $\bf{C_1}$ and $\bf{C_2}$ in Figure \ref{fig:intro-example}. \textsc{CCCT} can only yield binary outputs. 
For other neural models, predicting threshold is 0.5. 
}
\label{tab:caseres}
\end{table}

\begin{figure}[t]
\setlength{\belowcaptionskip}{-3mm}
\begin{center}
\includegraphics[width=0.4\textwidth]{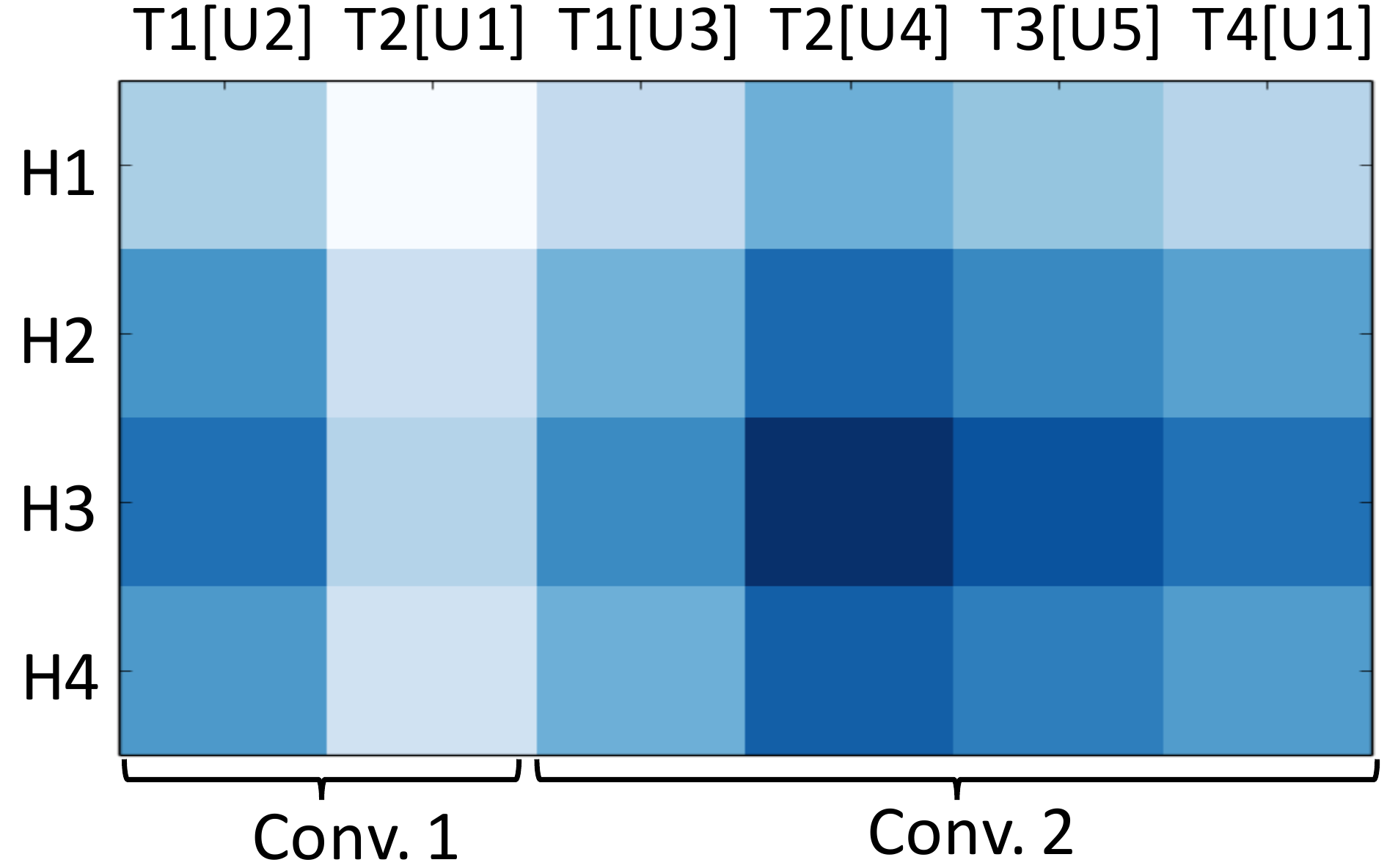}
\captionsetup{font=10pt}
\caption{
Attention output of model \textsc{BiLSTM+BiA} for the two sample conversations in Figure~\ref{fig:intro-example}. 
}\label{fig:attention}
\end{center}
\end{figure}

We further visualize the attention weights output by \textsc{BiLSTM+BiA}'s bi-attention mechanism with a heatmap in Figure \ref{fig:attention}. As can be seen, it assigns higher attention values to turns $\bf{T_2}$ and $\bf{T_3}$ in conversation $\bf{C_2}$, due to their topical similarity with user ${\bf U_1}$'s interests, i.e. movies, as inferred from their previous messages about \textit{Let Me In}. The attention weights then guide the final prediction for higher chance of re-entry to $\bf{C_2}$ rather than $\bf{C_1}$.


\subsection{Comparing with Humans}\label{ssec:human}
We are also interested in how human performs for the first re-entry prediction task, in order to find out how challenging such a task is. To achieve this, we design a human evaluation. 
Concretely, from each dataset, we randomly sample $50$ users who have been involved in at least $4$ conversations, with both re-entry and non re-entry behaviors exhibited. 
Then for each user $u$, we construct paired samples based on randomly selected conversations $c_1$ and $c_2$, where $u$ re-engage in one but not the other. 
The rest of the conversations that $u$ participated in are collected as their user history. 
Then, we invite two humans who are fluent speakers of English, to predict which conversation user $u$ will re-engage, after reading the context up to user's first participation in the paired conversations $c_1$ and $c_2$. 
They are requested to make a second prediction after reading user's chatting history. 

\begin{table}[t]
\begin{center}
\setlength{\tabcolsep}{4.0mm}
\fontsize{10}{11}\selectfont
\begin{tabular}{|l|c|c|}
\hline 
\bf{Predictor}&\bf{Twitter}&\bf{Reddit}\\
\hline
Human 1&26 ({\bf 29})&30 ({\bf 30})\\
Human 2&25 ({\bf 28})&28 ({\bf 29})\\
\textsc{BiLSTM+BiA}&35&33\\
\hline
\end{tabular} 
\end{center}
\captionsetup{font=10pt}
\caption{
Numbers of correct predictions made by humans, reading conversation context only and further seeing users' chatting history (boldfaced numbers), compared to the results of our best model in same setting. A random guess gives $25$ (out of $50$ pairs). 
}
\label{tab:he}
\end{table}

Humans' prediction performance is shown in Table~\ref{tab:he} along with \textsc{BiLSTM+BiA} model's output on the same data. 
As can be seen, humans can only give marginally better predictions than a random guess, i.e., 25 out of 50 pairs. Their performance improves after reading the user's previous posts, however, still falls behind our model's predictions. This indicates the ability of our model to learn from large-scaled data and align users' interests with conversation content. 
In addition, we notice that humans yield better performance on Reddit conversations than Twitter. It might be due to the fact that Reddit conversations are more focused, and it is easier for humans to identify the discussion points. While for Twitter discussions, the informal language usage further hinders humans' judgment.

\section{Conclusion}

We study the joint effects of conversation context and user chatting history for re-entry prediction. 
A novel neural framework is proposed for learning the interactions between two source of information.
Experimental results on two large-scale datasets from Twitter and Reddit show that our model with bi-attention yields better performance than the previous state of the art. 
Further discussions show that the model learns meaningful representations from conversation context and user history and hence exhibits consistent better performance given varying lengths of context or history. 
We also conduct a human study on the first re-entry prediction task. 
Our proposed model is observed to outperform humans, benefiting from its effective learning from large-scaled data.
\section*{Acknowledgements}
This work is partly supported by HK RGC GRF (14232816, 14209416, 14204118), NSFC (61877020). Lu Wang is supported in part by National Science Foundation through Grants IIS-1566382 and IIS-1813341. We thank the three anonymous reviewers for the insightful suggestions on various aspects of this work.

\bibliography{acl2019}
\bibliographystyle{acl_natbib}

\appendix

\end{document}